\documentclass[11pt,letterpaper]{article}
\usepackage{naaclhlt2016}
\usepackage{times}
\usepackage{latexsym}

\usepackage{url}
\usepackage{graphicx}
\usepackage{amsmath}
\usepackage{amsfonts}
\usepackage{amssymb}
\usepackage{dsfont}
\usepackage[english]{babel}
\usepackage{color}
\usepackage{paralist}
\usepackage{booktabs}

\def\ignore#1{}

\naaclfinalcopy % Uncomment this line for the final submission
\def\naaclpaperid{331} %  Enter the naacl Paper ID here

\title{Learning Natural Language Inference with LSTM}

\author{Shuohang Wang\\
		School of Information Systems\\
		Singapore Management University\\
		\texttt{shwang.2014@phdis.smu.edu.sg}
	\And 
	Jing Jiang\\
	School of Information Systems\\
	Singapore Management University\\
	\texttt{jingjiang@smu.edu.sg}}

\date{}

\begin{document}

\maketitle

\begin{abstract}
Natural language inference (NLI) is a fundamentally important task in natural language processing that has many applications.
The recently released Stanford Natural Language Inference (SNLI) corpus has made it possible to develop and evaluate learning-centered methods such as deep neural networks for natural language inference (NLI).
In this paper, we propose a special long short-term memory (LSTM) architecture for NLI.
Our model builds on top of a recently proposed neural attention model for NLI but is based on a significantly different idea.
Instead of deriving sentence embeddings for the premise and the hypothesis to be used for classification, 
our solution uses a match-LSTM to perform word-by-word matching of the hypothesis with the premise.
This LSTM is able to place more emphasis on important word-level matching results.
In particular, we observe that this LSTM remembers important mismatches that are critical for predicting the contradiction or the neutral relationship label.
On the SNLI corpus, our model achieves an accuracy of 86.1\%, outperforming the state of the art.
% Our experiments on the SNLI corpus show that our model outperforms the state of the art, achieving an accuracy of 86.1\% on the test data.
\end{abstract}
\section{Introduction}

Natural language inference (NLI) is the problem of determining whether from a premise sentence $P$ one can infer another hypothesis sentence $H$~\cite{maccartney:09}.
% Here $P$ is called the \emph{premise} and $H$ the \emph{hypothesis}.
NLI is a fundamentally important problem that has applications in many tasks including question answering, semantic search and automatic text summarization.
There has been much interest in NLI in the past decade, especially surrounding the PASCAL Recognizing Textual Entailment (RTE) Challenge~\cite{dagan:pascal05}. 
Existing solutions to NLI range from shallow approaches based on lexical similarities~\cite{glickman:pascal05} to advanced methods that consider syntax~\cite{mehdad:tac09}, perform explicit sentence alignment~\cite{maccartney:emnlp08} or use formal logic~\cite{clark:tac09}.

Recently, \newcite{bowman:emnlp15} released the Stanford Natural Language Inference (SNLI) corpus for the purpose of encouraging more learning-centered approaches to NLI.
This corpus contains around 570K sentence pairs with three labels: \emph{entailment}, \emph{contradiction} and \emph{neutral}.
The size of the corpus makes it now feasible to train deep neural network models, which typically require a large amount of training data.
\newcite{bowman:emnlp15} tested a straightforward architecture of deep neural networks for NLI.
In their architecture, the premise and the hypothesis are each represented by a sentence embedding vector.
The two vectors are then fed into a multi-layer neural network to train a classifier.
\newcite{bowman:emnlp15} achieved an accuracy of 77.6\% when long short-term memory (LSTM) networks were used to obtain the sentence embeddings.

A more recent work by \newcite{rocktaschel:iclr16} improved the performance by applying a neural attention model.
While their basic architecture is still based on sentence embeddings for the premise and the hypothesis, a key difference is that the embedding of the premise takes into consideration the alignment between the premise and the hypothesis.
This so-called \emph{attention-weighted} representation of the premise was shown to help push the accuracy to 83.5\% on the SNLI corpus.

A limitation of the aforementioned two models is that they reduce both the premise and the hypothesis to a single embedding vector before matching them; i.e., in the end, they use two embedding vectors to perform sentence-level matching.
However, not all word or phrase-level matching results are equally important.
For example, the matching between stop words in the two sentences is not likely to contribute much to the final prediction.
Also, for a hypothesis to \emph{contradict} a premise, a single word or phrase-level mismatch (e.g., a mismatch of the subjects of the two sentences) may be sufficient and other matching results are less important, but this intuition is hard to be captured if we directly match two sentence embeddings.

In this paper, we propose a new LSTM-based architecture for learning natural language inference.
Different from previous models, our prediction is not based on whole sentence embeddings of the premise and the hypothesis.
Instead, we use an LSTM to perform \emph{word-by-word} matching of the hypothesis with the premise.
Our LSTM sequentially processes the hypothesis, and at each position, it tries to match the current word in the hypothesis with an attention-weighted representation of the premise.
Matching results that are critical for the final prediction will be ``remembered'' by the LSTM while less important matching results will be ``forgotten.''
We refer to this architecture a match-LSTM, or \emph{m}LSTM for short.

Experiments show that our \emph{m}LSTM model achieves an accuracy of 86.1\% on the SNLI corpus, outperforming the state of the art.
Furthermore, through further analyses of the learned parameters, we show that the \emph{m}LSTM architecture can indeed pick up the more important word-level matching results that need to be remembered for the final prediction.
In particular, we observe that good word-level matching results are generally ``forgotten'' but important mismatches, which often indicate a \emph{contradiction} or a \emph{neutral} relationship, tend to be ``remembered.''

Our code is available online\footnote{\url{https://github.com/shuohangwang/SeqMatchSeq}}. 

\section{Model}

In this section, we first review LSTM.
We then review the word-by-word attention model by \newcite{rocktaschel:iclr16}, which is their best performing model.
Finally we present our \emph{m}LSTM architecture for natural language inference.

\subsection{Background}

\noindent \textbf{LSTM:} Let us first briefly review LSTM~\cite{hochreiter:nc97}.
LSTM is a special form of recurrent neural networks (RNNs), which process sequence data.
LSTM uses a few gate vectors at each position to control the passing of information along the sequence and thus improves the modeling of long-range dependencies.
While there are different variations of LSTMs, here we present the one adopted by \newcite{rocktaschel:iclr16}.
Specifically, let us use $\mathbf{X} = (\mathbf{x}_1, \mathbf{x}_2, \ldots, \mathbf{x}_N)$ to denote an input sequence, where  $\mathbf{x}_k \in \mathbb{R}^l$ ($1 \leq k \leq N$).
At each position $k$, there is a set of internal vectors, including an input gate $\mathbf{i}_k$, a forget gate $\mathbf{f}_k$, an output gate $\mathbf{o}_k$ and a memory cell $\mathbf{c}_k$.
All these vectors are used together to generate a $d$-dimensional hidden state $\mathbf{h}_k$ as follows:

\small
\begin{eqnarray}
\mathbf{i}_k & = & \sigma(\mathbf{W}^\text{i} \mathbf{x}_k + \mathbf{V}^\text{i} \mathbf{h}_{k-1} + \mathbf{b}^\text{i}), \nonumber \\
\mathbf{f}_k & = & \sigma(\mathbf{W}^\text{f} \mathbf{x}_k + \mathbf{V}^\text{f} \mathbf{h}_{k-1} + \mathbf{b}^\text{f}), \nonumber \\
\mathbf{o}_k & = & \sigma(\mathbf{W}^\text{o} \mathbf{x}_k + \mathbf{V}^\text{o} \mathbf{h}_{k-1} + \mathbf{b}^\text{o}), \nonumber \\
\mathbf{c}_k & = & \mathbf{f}_k \odot \mathbf{c}_{k-1} + \mathbf{i}_k \odot \tanh(\mathbf{W}^\text{c} \mathbf{x}_k + \mathbf{V}^\text{c} \mathbf{h}_{k-1} + \mathbf{b}^\text{c}), \nonumber \\
\mathbf{h}_k & = & \mathbf{o}_k \odot \tanh(\mathbf{c}_k),
\end{eqnarray}
\normalsize
where $\sigma$ is the sigmoid function, $\odot$ is the element-wise multiplication of two vectors, and all $\mathbf{W}^\text{*} \in \mathbb{R}^{d \times l}$,$\mathbf{V}^\text{*} \in \mathbb{R}^{d \times d}$ and $\mathbf{b}^\text{*} \in \mathbb{R}^d$ are weight matrices and vectors to be learned.

\noindent \textbf{Neural Attention Model:} For the natural language inference task, we have two sentences $\mathbf{X}^\text{s} = (\mathbf{x}^\text{s}_1, \mathbf{x}^\text{s}_2, \ldots, \mathbf{x}^\text{s}_M)$ and $\mathbf{X}^\text{t} = (\mathbf{x}^\text{t}_1, \mathbf{x}^\text{t}_2, \ldots, \mathbf{x}^\text{t}_N)$, where $\mathbf{X}^\text{s}$ is the premise and $\mathbf{X}^\text{t}$ is the hypothesis.
Here each $\mathbf{x}$ is an embedding vector of the corresponding word.
The goal is to predict a label $y$ that indicates the relationship between $\mathbf{X}^\text{s}$ and $\mathbf{X}^\text{t}$.
In this paper, we assume $y$ is one of \emph{entailment}, \emph{contradiction} and \emph{neutral}.

\newcite{rocktaschel:iclr16} first used two LSTMs to process the premise and the hypothesis, respectively, but initialized the second LSTM (for the hypothesis) with the last cell state of the first LSTM (for the premise).
Let us use $\mathbf{h}^\text{s}_j$ and $\mathbf{h}^\text{t}_k$ to denote the resulting hidden states corresponding to $\mathbf{x}^\text{s}_j$ and $\mathbf{x}^\text{t}_k$, respectively.
The main idea of the word-by-word attention model by \newcite{rocktaschel:iclr16} is to introduce a series of attention-weighted combinations of the hidden states of the premise, where each combination is for a particular word in the hypothesis.
Let us use $\mathbf{a}_k$ to denote such an \emph{attention vector} for word $\mathbf{x}^\text{t}_k$ in the hypothesis.
Specifically, $\mathbf{a}_k$ is defined as follows\footnote{We present the word-by-word attention model by \newcite{rocktaschel:iclr16} in a different way but the underlying model is the same. 
Our $\mathbf{h}^\text{a}_k$ is their $\mathbf{r}_t$,  our $\mathbf{H}^\text{s}$ (all of $\mathbf{h}^\text{s}_j$) is their $\mathbf{Y}$, our $\mathbf{h}^\text{t}_k$ is their $\mathbf{h}_\text{t}$, and our $\alpha_k$ is their $\alpha_t$.
%$\mathbf{w}^e$ to their $\mathbf{w}$, $\mathbf{W}^s$ to their $\mathbf{W}^y$, $\mathbf{W}^t$ to their $\mathbf{W}^h$, $\mathbf{W}^a$ to their $\mathbf{W}^r$,$\mathbf{V}^a$ to their $\mathbf{W}^t$.
Our presentation is close to the one by \newcite{bahdanau:icrl15}, with our attention vectors $\mathbf{a}$ corresponding to the context vectors $\mathbf{c}$ in their paper. }:

\small
\begin{eqnarray}
\label{eqn:a_k}
\mathbf{a}_k & = & \sum_{j=1}^M \alpha_{kj} \mathbf{h}^\text{s}_j,
\end{eqnarray}
\normalsize
where $\alpha_{kj}$ is an attention weight that encodes the degree to which $\mathbf{x}^\text{t}_k$ in the hypothesis is aligned with $\mathbf{x}^\text{s}_j$ in the premise.
The attention weight $\alpha_{kj}$ is generated in the following way:

\small
\begin{eqnarray}
\alpha_{kj} & = & \frac{\exp(e_{kj})}{\sum_{j'} \exp(e_{kj'})},
\end{eqnarray}
\normalsize
where 

\small
\begin{equation}
\label{eqn:e_kj}
e_{kj} = \mathbf{w}^\text{e} \cdot \tanh(\mathbf{W}^\text{s} \mathbf{h}^\text{s}_j + \mathbf{W}^\text{t} \mathbf{h}^\text{t}_k + \mathbf{W}^\text{a} \mathbf{h}^\text{a}_{k-1}).
\end{equation}
\normalsize
Here $\cdot$ is the dot-product between two vectors, the vector $\mathbf{w}^\text{e} \in \mathbb{R}^d$ and all matrices $\mathbf{W}^\text{*} \in \mathbb{R}^{d \times d}$ contain weights to be learned, and $\mathbf{h}^\text{a}_{k-1}$ is another hidden state which we will explain below.

The attention-weighted premise $\mathbf{a}_k$ essentially tries to model the relevant parts in the premise with respect to $\mathbf{x}^\text{t}_k$, i.e., the $k^{\mathrm{th}}$ word in the hypothesis.
\newcite{rocktaschel:iclr16} further built an RNN model over $\{\mathbf{a}_k\}_{k=1}^N$ by defining the following hidden states:

\small
\begin{eqnarray}
\mathbf{h}^\text{a}_k & = & \mathbf{a}_k + \tanh(\mathbf{V}^\text{a} \mathbf{h}^\text{a}_{k-1}),
\end{eqnarray}
\normalsize
where $\mathbf{V}^\text{a} \in \mathbb{R}^{d \times d}$ is a weight matrix to be learned.
We can see that the last $\mathbf{h}^\text{a}_N$ aggregates all the previous $\mathbf{a}_k$ and can be seen as an attention-weighted representation of the whole premise.
\newcite{rocktaschel:iclr16} then used this $\mathbf{h}^\text{a}_N$, which represents the whole premise, together with $\mathbf{h}^\text{t}_N$, which can be approximately regarded as an aggregated representation of the hypothesis\footnote{Strictly speaking, in the model by \newcite{rocktaschel:iclr16}, $\mathbf{h}^\text{t}_N$ encodes both the premise and the hypothesis because the two sentences are chained. But $\mathbf{h}^\text{t}_N$ places a higher emphasis on the hypothesis given the nature of RNNs.},
to predict the label $y$.

\subsection{Our Model}

Although the neural attention model by \newcite{rocktaschel:iclr16} achieved better results than \newcite{bowman:emnlp15}, we see two limitations.
First, the model still uses a single vector representation of the premise, namely $\mathbf{h}^\text{a}_N$, to match the entire hypothesis.
We speculate that if we instead use each of the attention-weighted representations of the premise for matching, i.e., use $\mathbf{a}_k$ at position $k$ to match the hidden state $\mathbf{h}^\text{t}_k$ of the hypothesis while we go through the hypothesis, we could achieve better matching results.
This can be done using an RNN which at each position takes in both $\mathbf{a}_k$ and $\mathbf{h}^\text{t}_k$ as its input and determines how well the overall matching of the two sentences is up to the current position.
In the end the RNN will produce a single vector representing the matching of the two entire sentences.

The second limitation is that the model by \newcite{rocktaschel:iclr16} does not explicitly allow us to place more emphasis on the more important matching results between the premise and the hypothesis and down-weight the less critical ones.
For example, matching of stop words is presumably less important than matching of content words.
Also, some matching results may be particularly critical for making the final prediction and thus should be remembered.
For example, consider the premise ``\emph{A dog jumping for a Frisbee in the snow.}'' and the hypothesis ``\emph{A cat washes his face and whiskers with his front paw.}''
When we sequentially process the hypothesis, once we see that the subject of the hypothesis \emph{cat} does not match the subject of the premise \emph{dog}, we have a high probability to believe that there is a contradiction.
So this mismatch should be remembered.

\begin{figure}[t]
\begin{center}
\includegraphics[width=3.0in]{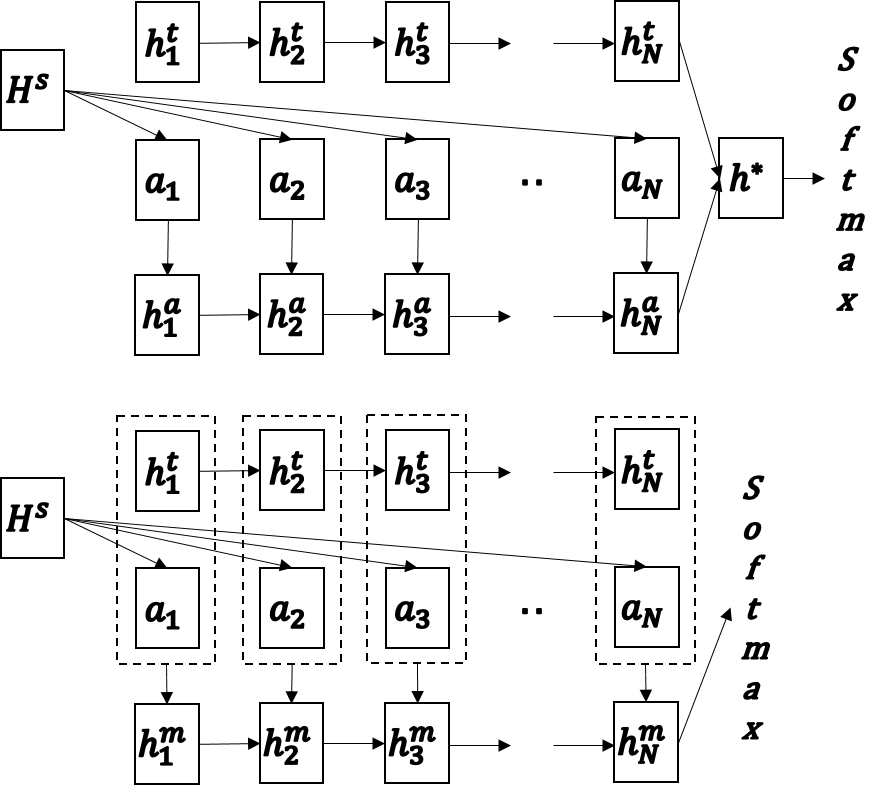}
\end{center}
\caption{The top figure depicts the model by Rockt{\"a}schel et al. (2016)
and the bottom figure depicts our model.
Here $\mathbf{H}^\text{s}$ represents all the hidden states $\mathbf{h}^\text{s}_j$.
Note that in the top model 
each $\mathbf{h}^\text{a}_k$ represents a weighted version of the premise only, while in our model, each $\mathbf{h}^\text{m}_k$ represents the matching between the premise and the hypothesis up to position $k$.}
\label{fig:model}
\end{figure}

Based on the two observations above, we propose to use an LSTM to sequentially match the two sentences.
At each position the LSTM takes in both $\mathbf{a}_k$ and $\mathbf{h}^\text{t}_k$ as its input.
% The LSTM is expected to remember the more important matches between the two sentences for predicting the final label $y$ and forget the less important ones.
Figure~\ref{fig:model} gives an overview of our model in contrast to the model by \newcite{rocktaschel:iclr16}.

Specifically, our model works as follows.
First, similar to \newcite{rocktaschel:iclr16}, we process the premise and the hypothesis using two LSTMs, but we do not feed the last cell state of the premise to the LSTM of the hypothesis.
This is because we do not need the LSTM for the hypothesis to encode any knowledge about the premise but we will match the premise with the hypothesis using the hidden states of the two LSTMs.
Again, we use $\mathbf{h}^\text{s}_j$ and $\mathbf{h}^\text{t}_k$ to represent these hidden states.

Next, we generate the attention vectors $\mathbf{a}_k$ similarly to Eqn~(\ref{eqn:a_k}).
However, Eqn~(\ref{eqn:e_kj}) will be replaced by the following equation:

\small
\begin{equation}
e_{kj} = \mathbf{w}^\text{e} \cdot \tanh(\mathbf{W}^\text{s} \mathbf{h}^\text{s}_j + \mathbf{W}^\text{t} \mathbf{h}^\text{t}_k + \mathbf{W}^\text{m} \mathbf{h}^\text{m}_{k-1}).
\end{equation}
\normalsize
The only difference here is that we use a hidden state $\mathbf{h}^\text{m}$ instead of $\mathbf{h}^\text{a}$, and the way we define $\mathbf{h}^\text{m}$ is very different from the definition of $\mathbf{h}^\text{a}$.

Our $\mathbf{h}^\text{m}_k$ is the hidden state at position $k$ generated from our \emph{m}LSTM.
This LSTM models the \emph{matching} between the premise and the hypothesis.
Important matching results will be ``remembered'' by the LSTM while non-essential ones will be ``forgotten.''
We use the concatenation of $\mathbf{a}_k$, which is the attention-weighted version of the premise for the $k^{\mathrm{th}}$ word in the hypothesis, and $\mathbf{h}^\text{t}_k$, the hidden state for the $k^{\mathrm{th}}$ word itself, as input to the \emph{m}LSTM.

Specifically, let us define

\small
\begin{eqnarray}
\mathbf{m}_k & = & \begin{bmatrix}
\mathbf{a}_k \\
\mathbf{h}^\text{t}_k
\end{bmatrix}.
\end{eqnarray}
\normalsize
We then build the \emph{m}LSTM as follows:

\small
\begin{eqnarray}
\nonumber
\mathbf{i}^\text{m}_k & = & \sigma(\mathbf{W}^\text{mi} \mathbf{m}_k + \mathbf{V}^\text{mi} \mathbf{h}^\text{m}_{k-1} + \mathbf{b}^\text{mi}), \\
\nonumber
\mathbf{f}^\text{m}_k & = & \sigma(\mathbf{W}^\text{mf} \mathbf{m}_k + \mathbf{V}^\text{mf} \mathbf{h}^\text{m}_{k-1} + \mathbf{b}^\text{mf}), \\
\nonumber
\mathbf{o}^\text{m}_k & = & \sigma(\mathbf{W}^\text{mo} \mathbf{m}_k + \mathbf{V}^\text{mo} \mathbf{h}^\text{m}_{k-1} + \mathbf{b}^\text{mo}), \\
\nonumber
\mathbf{c}^\text{m}_k & = & \mathbf{f}^\text{m}_k \odot \mathbf{c}^\text{m}_{k-1} + \mathbf{i}^\text{m}_k \odot \tanh(\mathbf{W}^\text{mc} \mathbf{m}_k + \mathbf{V}^\text{mc} \mathbf{h}^\text{m}_{k-1} \\
\nonumber
& & + \mathbf{b}^\text{mc}), \\
\mathbf{h}^\text{m}_k & = & \mathbf{o}^\text{m}_k \odot \tanh(\mathbf{c}^\text{m}_k).
\end{eqnarray}
\normalsize
With this \emph{m}LSTM, finally we use only $\mathbf{h}^\text{m}_N$, the last hidden state, to predict the label $y$.

\subsection{Implementation Details}
\label{subsec:details}

Besides the difference of the LSTM architecture, we also introduce a few other changes from the model by \newcite{rocktaschel:iclr16}.
First, we insert a special word \emph{NULL} to the premise, and we allow words in the hypothesis to be aligned with this \emph{NULL}.
This is inspired by common practice in machine translation.
Specifically, we introduce a vector $\mathbf{h}^\text{s}_0$, which is fixed to be a vector of 0s of dimension $d$.
This $\mathbf{h}^\text{s}_0$ represents \emph{NULL} and is used with other $\mathbf{h}^\text{s}_j$ to derive the attention vectors $\{\mathbf{a}_k\}_{k=1}^N$.

Second, we use word embeddings trained from GloVe~\cite{pennington:emnlp14} instead of word2vec vectors.
The main reason is that GloVe covers more words in the SNLI corpus than word2vec\footnote{The SNLI corpus contains $~$37K unique tokens. Around 12.1K of them cannot be found in word2vec but only around 4.1K of them cannot be found in GloVe.}.

Third, for words which do not have pre-trained word embeddings, we take the average of the embeddings of all the words (in GloVe) surrounding the unseen word within a window size of 9 (4 on the left and 4 on the right) as an approximation of the embedding of this unseen word.
Then we do not update any word embedding when learning our model.
Although this is a very crude approximation, it reduces the number of parameters we need to update, and as it turns out, we can still achieve better performance than \newcite{rocktaschel:iclr16}.

\section{Experiments}

\begin{table*}[t]
	\centering
	\small
	\begin{tabular}{lllllll}
		\toprule
		Model                  & $d$   & $|\theta|_{\text{W+M}}$ & $|\theta|_{\text{M}}$ & Train & Dev  & Test \\ 
		\midrule
		LSTM [\newcite{bowman:emnlp15}]                  & 100 & 10M            & 221K         & 84.4  & -    & 77.6 \\
		Classifier [\newcite{bowman:emnlp15}]             & -   & -              & -            & 99.7  & -    & 78.2 \\ 
		\midrule
		LSTM shared [\newcite{rocktaschel:iclr16}]           & 159 & 3.9M           & 252K         & 84.4  & 83.0 & 81.4 \\
		Word-by-word attention [\newcite{rocktaschel:iclr16}] & 100 & 3.9M           & 252K         & 85.3  & 83.7 & 83.5 \\ \midrule
		Word-by-word attention (our implementation)  & 150 & 340K           & 340K         & 85.5  & 83.3 & 82.6 \\ 
		\emph{m}LSTM     & 150 &         544K       &     544K        & 91.0    & 86.2 & 85.7 \\
		\emph{m}LSTM with bi-LSTM sentence modeling     & 150 &         1.4M       &     1.4M        & 91.3    & 86.6 & 86.0 \\ 
		\emph{m}LSTM     & 300 &         1.9M       &     1.9M        & 92.0    & \textbf{86.9} & \textbf{86.1} \\
		\emph{m}LSTM with word embedding     & 300 &        1.3M        &            1.3M &  88.6   & 85.4 & 85.3 \\
		\bottomrule
	\end{tabular}
	\normalsize
	\caption{Experiment results in terms of accuracy.
		$d$ is the dimension of the hidden states.
		$|\theta|_{\text{W+M}}$ is the total number of parameters and $|\theta|_{\text{M}}$ is the number of parameters excluding the word embeddings.
		Note that the five models in the last section were implemented by us while the other results were taken directly from previous papers.
		Note also that for the five models in the last section, we do not update word embeddings so $|\theta|_{\text{W+M}}$  is the same as $|\theta|_{\text{M}}$.
		The three columns on the right are the accuracies of the trained models on the training data, the development data and the test data, respectively.} 
	\label{tab:results}
\end{table*}

% In this section, we present the evaluation of our model.
% We first present the main experiment results, comparing our model with the model by \newcite{rocktaschel:iclr16}.
% We then conduct some further analyses to understand how our \emph{m}LSTM model works in matching the premise and the hypothesis.

\subsection{Experiment Settings}

\noindent \textbf{Data:} We use the SNLI corpus to test the effectiveness of our model.
The original data set contains 570,152 sentence pairs, each labeled with one of the following relationships: \emph{entailment}, \emph{contradiction}, \emph{neutral} and \emph{--}, where \emph{--} indicates a lack of consensus from the human annotators.
We discard the sentence pairs labeled with \emph{--} and keep the remaining ones for our experiments.
In the end, we have 549,367 pairs for training, 9,842 pairs for development and 9,824 pairs for testing.
This follows the same data partition used by \newcite{bowman:emnlp15} in their experiments.
We perform three-class classification and use accuracy as our evaluation metric.

% We also use the SICK corpus~\cite{marelli:lrec14}, a benchmark data set commonly used for recognizing textual entailment, to check how well the model trained on the SNLI corpus can perform on a different data set.
% \textcolor{magenta}{The SICK corpus contains 10,000 sentence pairs, where each pair is annotated with a score indicating the relatedness of the two sentences and a label indicating their entailment relationship.
% Just like the SNLI corpus, there are three entailment labels: \emph{entailment}, \emph{contradiction} and \emph{neutral}.}

\noindent \textbf{Parameters:} We use the Adam method~\cite{kingma:arxiv14} with hyperparameters $\beta_1$ set to 0.9 and $\beta_2$ set to 0.999 for optimization.
The initial learning rate is set to be 0.001 with a decay ratio of 0.95 for each iteration.
The batch size is set to be 30.
We experiment with $d = 150$ and $d = 300$ where $d$ is the dimension of all the hidden states.
% of the LSTMs.

\noindent \textbf{Methods for comparison:} We mainly want to compare our model with the word-by-word attention model by \newcite{rocktaschel:iclr16} because this model achieved the state-of-the-art performance on the SNLI corpus.
To ensure fair comparison, besides comparing with the accuracy reported by \newcite{rocktaschel:iclr16}, we also re-implemented their model and report the performance of our implementation.
We also consider a few variations of our model.
Specifically, the following models are implemented and tested in our experiments:

\begin{compactitem}
	\item Word-by-word attention ($d = 150$): This is our implementation of the word-by-word attention model by \newcite{rocktaschel:iclr16}, where we set the dimension of the hidden states to 150.
	The differences between our implementation and the original implementation by \newcite{rocktaschel:iclr16} are the following: 
	(1) We also add a \emph{NULL} token to the premise for matching.
	(2) We do not feed the last cell state of the LSTM for the premise to the LSTM for the hypothesis, to keep it consistent with the implementation of our model.
	(3) For word representation, we also use the GloVe word embeddings and we do not update the word embeddings.
	For unseen words, we adopt the same strategy as described in Section~\ref{subsec:details}.
	
	\item \emph{m}LSTM ($d = 150$): This is our \emph{m}LSTM model with $d$ set to 150. 
	
	\item \emph{m}LSTM with bi-LSTM sentence modeling ($d = 150$): This is the same as the model above except that when we derive the hidden states $\mathbf{h}^\text{s}_j$ and $\mathbf{h}^\text{t}_k$ of the two sentences, we use bi-LSTMs~\cite{graves2012supervised} instead of LSTMs.
	We implement this model to see whether bi-LSTMs allow us to better align the sentences.
	
	\item \emph{m}LSTM ($d = 300$): This is our \emph{m}LSTM model with $d$ set to 300.
	
	\item \emph{m}LSTM with word embedding ($d = 300$): This is the same as the model above except that we directly use the word embedding vectors $\mathbf{x}^\text{s}_j$ and $\mathbf{x}^\text{t}_k$ instead of the hidden states $\mathbf{h}^\text{s}_j$ and $\mathbf{h}^\text{t}_k$ in our model.
	In this case, each attention vector $\mathbf{a}_k$ is a weighted sum of $\{\mathbf{x}^\text{s}_j\}_{j=1}^M$.
	We experiment with this setting because we hypothesize that the effectiveness of our model is largely related to the \emph{m}LSTM architecture rather than the use of LSTMs to process the original sentences.
	
\end{compactitem}

\begin{table}[]
	\centering
	\small
	\begin{tabular}{c|ccc}
		\toprule
		& \multicolumn{3}{c}{ground truth} \\
		prediction &  \emph{N}            & \emph{E}           & \emph{C}           \\ \midrule
		\emph{N}   & 2628    & 286  & 255  \\
		\emph{E}   & 340 & 3005   & 159  \\ 
		\emph{C}   & 250 & 77 & 2823 \\ \bottomrule
	\end{tabular}
	\normalsize
	\caption{The confusion matrix of the results by \emph{m}LSTM with $d = 300$. 
		\emph{N}, \emph{E} and \emph{C} correspond to \emph{neutral}, \emph{entailment} and \emph{contradiction}, respectively.}
	\label{tab:confusion_m}
\end{table}

\ignore{
\begin{table}[]
\centering
\scriptsize
\begin{tabular}{c|ccc}
\toprule
& \multicolumn{3}{c}{ground truth} \\
prediction &  \emph{N}            & \emph{E}           & \emph{C}           \\ \midrule
\emph{N}   & 2628 (81.7\%)    & 286 (8.5\%)  & 255 (7.9\%) \\
\emph{E}   & 340 (10.6\%) & 3005 (89.2\%)  & 159 (5.0\%) \\ 
\emph{C}   & 250 (7.8\%)  & 77 (2.3\%) & 2823 (87.2\%) \\ \bottomrule
\end{tabular}
\normalsize
\caption{The confusion matrix of the results by \emph{m}LSTM with $d = 300$. 
		\emph{N}, \emph{E} and \emph{C} correspond to \emph{neutral}, \emph{entailment} and \emph{contradiction}, respectively.}
\label{tab:confusion_m}
\end{table}
}

\ignore{
\begin{table}[]
\centering
\scriptsize
\begin{tabular}{ll}
\hline
Model & Test \\ \hline
LSTM [\newcite{bowman:emnlp15}] & 46.7 \\ 
Word-by-word attention (our implementation)     & 49.1 \\ 
mlstm     & 55.2 \\ \hline
\end{tabular}
\normalsize
\caption{\textcolor{blue}{The results of testing on SICK data set by using the models trained only on SNLI data set.}}
\label{tab:sick}
\end{table}
}

\begin{table*}[t]
	\centering
	\small
	\begin{tabular}{llll}
		\hline
		& \textbf{ID} & \textbf{sentence} & \textbf{label} \\
		\hline
		Premise & & A dog jumping for a Frisbee in the snow. & \\
		\hline
		& Example 1 & An animal is outside in the cold weather, playing with a plastic toy. & \emph{entailment} \\
		Hypothesis & Example 2 & A cat washed his face and whiskers with his front paw. & \emph{contradiction} \\
		& Example 3 & A pet is enjoying a game of fetch with his owner. & \emph{neutral}   \\
		\hline
	\end{tabular}
	\normalsize
	\caption{Three examples of sentence pairs with different relationship labels.
		The second hypothesis is a contradiction because it mentions a completely different event.
		The third hypothesis is neutral to the premise because the phrase ``with his owner'' cannot be inferred from the premise.}
	\label{tab:examples}
\end{table*}

\subsection{Main Results}

Table~\ref{tab:results} compares the performance of the various models we tested together with some previously reported results.

We have the following observations:
(1) First of all, we can see that when we set $d$ to 300, our model achieves an accuracy of 86.1\% on the test data, which to the best of our knowledge is the highest on this data set.
(2) If we compare our \emph{m}LSTM model with our implementation of the word-by-word attention model by \newcite{rocktaschel:iclr16} under the same setting with $d = 150$, we can see that our performance on the test data (85.7\%) is higher than that of their model (82.6\%).
We also tested statistical significance and found the improvement to be statistically significant at the 0.001 level.
(3) The performance of \emph{m}LSTM with bi-LSTM sentence modeling compared with the model with standard LSTM sentence modeling when $d$ is set to 150 shows that using bi-LSTM to process the original sentences helps (86.0\% vs. 85.7\% on the test data), but the difference is small and the complexity of bi-LSTM is much higher than LSTM.
Therefore when we increased $d$ to 300 we did not experiment with bi-LSTM sentence modeling.
(4) Interestingly, when we experimented with the \emph{m}LSTM model using the pre-trained word embeddings instead of LSTM-generated hidden states as initial representations of the premise and the hypothesis, we were able to achieve an accuracy of 85.3\% on the test data, which is still better than previously reported state of the art.
This suggests that the \emph{m}LSTM architecture coupled with the attention model works well, regardless of whether or not we use LSTM to process the original sentences.

Because the NLI task is a three-way classification problem, to better understand the errors, we also show the confusion matrix of the results obtained by our \emph{m}LSTM model with $d = 300$ in Table~\ref{tab:confusion_m}.
We can see that there is more confusion between \emph{neutral} and \emph{entailment} and between \emph{neutral} and \emph{contradiction} than between \emph{entailment} and \emph{contradiction}.
This shows that \emph{neutral} is relatively hard to capture.

% We further analyse our model \emph{m}LSTM through the confusion matrix shown in Table~\ref{tab:confusion_m}. It can be seen that the \emph{Neural} relationship between sentences is more likely to be confused than the others. This may be caused by mis-alignments for the words in hypothesis that can be neither entailed nor contradicted by premise.

% \textcolor{blue}{Besides, we also test the adaptation of our model through Table~\ref{tab:sick} which shows the results tested on Text Entailment task of SICK \cite{marelli2014semeval} by using the model trained only on SNLI dataset. We didn't use any transfer learning techniques, and our model get a relatively better accuracy 55.2\% on the test data. }

\subsection{Further Analyses}

\begin{figure*}[!ht]
\centering
\includegraphics[width=6in]{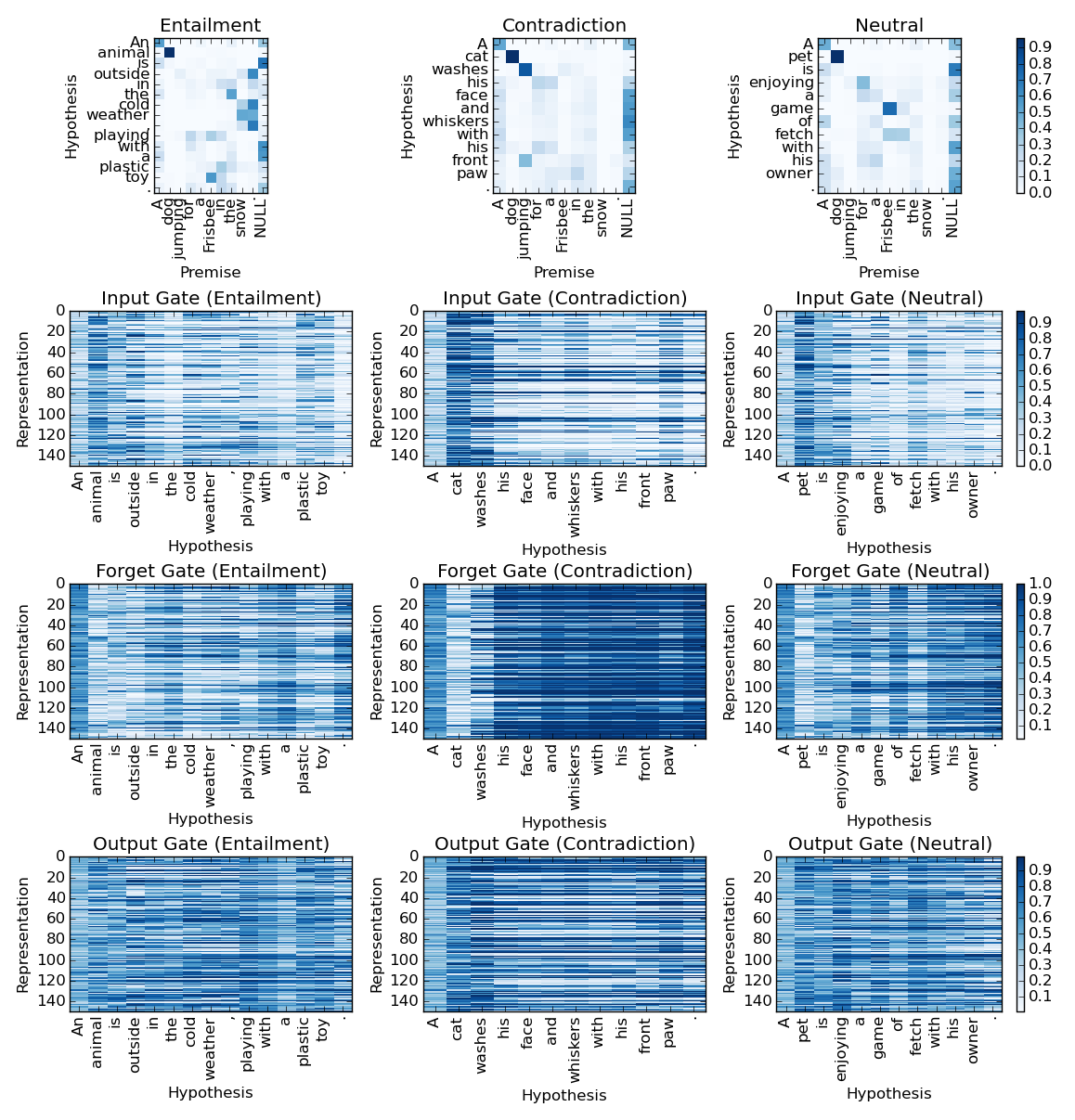}
\caption{The alignment weights and the gate vectors of the three examples.}
\label{fig:qualitative}
\end{figure*}

To obtain a better understanding of how our proposed model actually performs the matching between a premise and a hypothesis, we further conduct the following analyses.
First, we look at the learned word-by-word alignment weights $\alpha_{kj}$ to check whether the soft alignment makes sense.
This is the same as what was done by \newcite{rocktaschel:iclr16}.
We then look at the values of the various gate vectors of the {\emph{m}LSTM}.
By looking at these values, we aim to check (1) whether the model is able to differentiate between more important and less important word-level matching results, and (2) whether the model forgets certain matching results and remembers certain other ones.

To conduct the analyses, we choose three examples and display the various learned parameter values.
These three sentence pairs share the same premise but have different hypotheses and different relationship labels.
They are given in Table~\ref{tab:examples}.
The values of the alignment weights and the gate vectors are plotted in Figure~\ref{fig:qualitative}.

Besides using the three examples, we will also give some overall statistics of the parameter values to confirm our observations with the three examples.

\subsubsection*{Word Alignment}

First, let us look at the top-most plots of Figure~\ref{fig:qualitative}.
These plots show the alignment weights $\alpha_{kj}$ between the hypothesis and the premise, where a darker color corresponds to a larger value of $\alpha_{kj}$.
Recall that $\alpha_{kj}$ is the degree to which the $k^{\mathrm{th}}$ word in the hypothesis is aligned with the $j^{\mathrm{th}}$ word in the premise.
Also recall that the weights $\alpha_{kj}$ are configured such that for the same $k$ all the $\alpha_{kj}$ add up to 1.
This means the weights in the same row in these plots add up to 1.
From the three plots we can see that the alignment weights generally make sense.
For example, in Example 1, ``animal'' is strongly aligned with ``dog'' and ``toy'' aligned with ``Frisbee.''
The phrase ``cold weather'' is aligned with ``snow.''
In Example 3, we also see that ``pet'' is strongly aligned with ``dog'' and ``game'' aligned with ``Frisbee.''

In Example 2, ``cat'' is strongly aligned with ``dog'' and ``washes'' is aligned with ``jumping.''
It may appear that these matching results are wrong.
However, ``dog'' is likely the best match for ``cat'' among all the words in the premise, and as we will show later, this match between ``cat'' and ``dog'' is actually a strong indication of a contradiction between the two sentences.
The same explanation applies to the match between ``washes'' and ``jumping.''

We also observe that some words are aligned with the \emph{NULL} token we inserted.
For example, the word ``is'' in the hypothesis in Example 1 does not correspond to any word in the premise and is therefore aligned with \emph{NULL}.
The words ``face'' and ``whiskers'' in Example 2 and ``owner'' in Example 3 are also aligned with \emph{NULL}.
Intuitively, if some important content words in the hypothesis 
are aligned with \emph{NULL}, it is more likely that the relationship label is either contradiction or neutral.

\subsubsection*{Values of Gate Vectors}

Next, let us look at the values of the learned gate vectors of our \emph{m}LSTM for the three examples.
We show these values under the setting where $d$ is set to 150.
Each row of these plots corresponds to one of the 150 dimensions.
Again, a darker color indicates a higher value.

An input gate controls whether the input at the current position should be used in deriving the final hidden state of the current position.
From the three plots of the input gates, we can observe that generally for stop words such as prepositions and articles the input gates have lower values, suggesting that the matching of these words is less important.
On the other hand, content words such as nouns and verbs tend to have higher values of the input gates, which also makes sense because these words are generally more important for determining the final relationship label. 

To further verify the observation above, we compute the average input gate values for stop words and the other content words. 
We find that the former has an average value of 0.287 with a standard deviation of 0.084 while the latter has an average value of 0.347 with a standard deviation of 0.116. 
This shows that indeed generally stop words have lower input gate values.
Interestingly, we also find that some stop words may have higher input gate values if they are critical for the classification task.
For example, the negation word ``not'' has an average input gate value of 0.444 with a standard deviation of 0.104.

Overall, the values of the input gates confirm that the \emph{m}LSTM helps differentiate the more important word-level matching results from the less important ones. 

Next, let us look at the forget gates.
Recall that a forget gate controls the importance of the \emph{previous} cell state in deriving the final hidden state of the current position.
Higher values of a forget gate indicate that we need to remember the previous cell state and pass it on whereas lower values indicate that we should probably forget the previous cell.
From the three plots of the forget gates, we can see that overall the colors are the lightest for Example 1, which is an \emph{entailment}.
This suggests that when the hypothesis is an entailment of the premise, the \emph{m}LSTM tends to forget the previous matching results.
On the other hand, for Example 2 and Example 3, which are \emph{contradiction} and \emph{neutral}, we see generally darker colors.
In particular, in Example 2, we can see that the colors are consistently dark starting from the word ``his'' in the hypothesis until the end.
We believe the explanation is that after the \emph{m}LSTM processes the first three words of the hypothesis, ``A cat washes,'' it sees that the matching between ``cat'' and ``dog'' and between ``washes'' and ``jumping'' is a strong indication of a contradiction, and therefore these matching results need to be remembered until the end of the \emph{m}LSTM for the final prediction.

We have also checked the forget gates of the other sentence pairs in the test data by computing the average forget gate values and the standard deviations for \emph{entailment}, \emph{neutral} and \emph{contradiction}, respectively.
We find that the values are 0.446$\pm$0.123, 0.507$\pm$0.148 and 0.536$\pm$0.170, respectively.
For \emph{contradiction} and \emph{neutral}, the forget gates start to have higher values from certain positions of the hypotheses.

Based on the observations above, we hypothesize that the way the \emph{m}LSTM works is as follows. It remembers important mismatches, which are useful for predicting the \emph{contradiction} or the \emph{neutral} relationship, and forgets good matching results.
At the end of the \emph{m}LSTM, if no important mismatch is remembered, the final classifier will likely predict \emph{entailment} by default.
Otherwise, depending on the kind of mismatch remembered, the classifier will predict either \emph{contradiction} or \emph{neutral}.

% It is also interesting to point out that the values of the input gates seem to have a negative correlation with the values of the forget gates.
% In other words, at each position of the hypothesis, if the input gate has high values, then the forget gate tends to have low values, and vice versa.

For the output gates, we are not able to draw any important conclusion except that the output gates seem to be positively correlated with the input gates but they tend to be darker than the input gates.

\section{Related Work}

There has been much work on natural language inference.
Shallow methods rely mostly on lexical similarities but are shown to be robust.
For example, \newcite{bowman:emnlp15} experimented with a lexicalized classifier-based method, which only uses lexical information 
% to extract features used by a classifier, and the method 
and achieves an accuracy of 78.2\% on the SNLI corpus.
More advanced methods use syntactic structures of the sentences to help matching them.
For example, \newcite{mehdad:tac09} applied syntactic-semantic tree kernels for recognizing textual entailment.
Because inference is essentially a logic problem, methods based on formal logic~\cite{clark:tac09} or natural logic~\cite{maccartney:09} have also been proposed.
A comprehensive review on existing work 
% on natural language inference 
can be found in the book by \newcite{dagan:bk13}.

The work most relevant to ours is the recently proposed neural attention model-based method by \newcite{rocktaschel:iclr16}, which we have detailed in previous sections.
Neural attention models have recently been applied to some natural language processing tasks including machine translation~\cite{bahdanau:icrl15}, abstractive summarization~\cite{rush2015neural} and question answering~\cite{hermann2015teaching}.
\newcite{rocktaschel:iclr16} showed that the neural attention model could help derive a better representation of the premise to be used to match the hypothesis, whereas in our work we also use it to derive representations of the premise that are used to sequentially match the words in the hypothesis.

The SNLI corpus is new and so far it has only been used in a few studies.
Besides the work by \newcite{bowman:emnlp15} themselves and by \newcite{rocktaschel:iclr16}, there are two other studies which used the SNLI corpus.
\newcite{vendrov2015order} used a \emph{Skip-Thought} model proposed by \newcite{kiros2015skip} to the NLI task and reported an accuracy of 81.5\% on the test data.
\newcite{mou2015recognizing} used tree-based CNN encoders to obtain sentence embeddings and achieved an accuracy of 82.1\%.

% Because this accuracy is lower than the best performance reported by \newcite{rocktaschel:arxiv15} and because our main focus was to examine the effectiveness of our match-LSTM compared with the model by \newcite{rocktaschel:arxiv15}, we did not include their study for comparison in our experiments.

\section{Conclusions and Future Work}

In this paper, we proposed a special LSTM architecture for the task of natural language inference.
Based on a recent work by \newcite{rocktaschel:iclr16}, we first used neural attention models to derive attention-weighted vector representations of the premise.
We then designed a match-LSTM that processes the hypothesis word by word while trying to match the hypothesis with the premise.
The last hidden state of this \emph{m}LSTM can be used for predicting the relationship between the premise and the hypothesis.
Experiments on the SNLI corpus showed that the \emph{m}LSTM model outperformed the state-of-the-art performance reported so far on this data set.
Moreover, closer analyses on the gate vectors revealed that our \emph{m}LSTM indeed remembers and passes on important matching results, which are typically mismatches that indicate a \emph{contradiction} or a \emph{neutral} relationship between the premise and the hypothesis.

With the large number of parameters to learn, an inevitable limitation of our model 
is that a large training data set is needed to learn good model parameters.
Indeed some preliminary experiments applying our \emph{m}LSTM to the SICK corpus~\cite{marelli:lrec14}, a smaller textual entailment benchmark data set, did not give very good results.
We believe that this is because our model learns everything from scratch except using the pre-trained word embeddings.
A future direction would be to incorporate other resources such as the paraphrase database~\cite{ganitkevitch:naacl13} into the learning process.

\bibliography{naaclhlt2016}
\bibliographystyle{naaclhlt2016}

\end{document}